\documentclass[10pt,twocolumn,letterpaper]{article}

\usepackage{cvpr}
\usepackage{times}
\usepackage{epsfig}
\usepackage{graphicx}
\usepackage{amsmath}
\usepackage{amssymb}
\usepackage{bm}
\usepackage{tabularx} %

\usepackage{float}
\usepackage{wrapfig}
\usepackage{subcaption}
\usepackage{capt-of}%
\usepackage{booktabs}
\usepackage{varwidth}
\usepackage{wrapfig}
\usepackage{makecell}

\usepackage{paralist}

\usepackage{lipsum}
\newcommand\blfootnote[1]{%
  \begingroup
  \renewcommand\thefootnote{}\footnote{#1}%
  \addtocounter{footnote}{-1}%
  \endgroup
}
\usepackage{multirow}
\usepackage[pagebackref=true,breaklinks=true,letterpaper=true,colorlinks,bookmarks=false]{hyperref}

\cvprfinalcopy %

\def\cvprPaperID{329} %

\ifcvprfinal\fi
\begin{document}

\title{Reducing Uncertainty in Undersampled MRI Reconstruction \\ with Active Acquisition}

\author{Zizhao Zhang$^{1,2,*}$ Adriana Romero$^2$ Matthew J. Muckley$^3$ Pascal Vincent$^2$ Lin Yang$^1$ Michal Drozdzal$^2$ \\
$^1$ University of Florida \; $^2$ Facebook AI Research \; $^3$ NYU School of Medicine }

\maketitle

\begin{abstract}
The goal of MRI reconstruction is to restore a high fidelity image from partially observed measurements. This partial view naturally induces reconstruction uncertainty that can only be reduced by acquiring additional measurements. In this paper, we present a novel method for MRI reconstruction that, at inference time, dynamically selects the measurements to take and iteratively refines the prediction in order to best reduce the reconstruction error and, thus, its uncertainty. We validate our method on a large scale knee MRI dataset, as well as on ImageNet. Results show that (1) our system successfully outperforms active acquisition baselines; (2) our uncertainty estimates correlate with error maps; and (3) our ResNet-based architecture surpasses standard pixel-to-pixel models in the task of MRI reconstruction. The proposed method not only shows high-quality reconstructions but also paves the road towards more applicable solutions for accelerating MRI. 
\end{abstract}

\section{Introduction}
\label{sec:intro}
Magnetic Resonance Imaging (MRI) is a commonly used scanning technique, which provides detailed images of organs and tissues within the human body.  The promises of MRI, when compared to computed tomography, are the superior soft tissue contrast and the lack of ionizing radiation \cite{zhu2018image}. However, its main drawback is the slow acquisition time; MRI examinations can take as long as an hour. The acquisition is performed sequentially in \emph{k-space} -- a 2D complex-valued space that can be linked to the 2D Fourier transform of the image -- at speed controlled by hardware and physiological constraints \cite{moratal2008k,schlemper2018deep}, causing uncomfortable examination experiences and high health care costs. Therefore, accelerating MRI is a critical medical imaging problem, with the potential of substantially improving both its accessibility and the patient experience.
\blfootnote{*Work done during internship at Facebook AI Research} 

\begin{figure}[t!]
    \centering
    \includegraphics[width=0.45\textwidth]{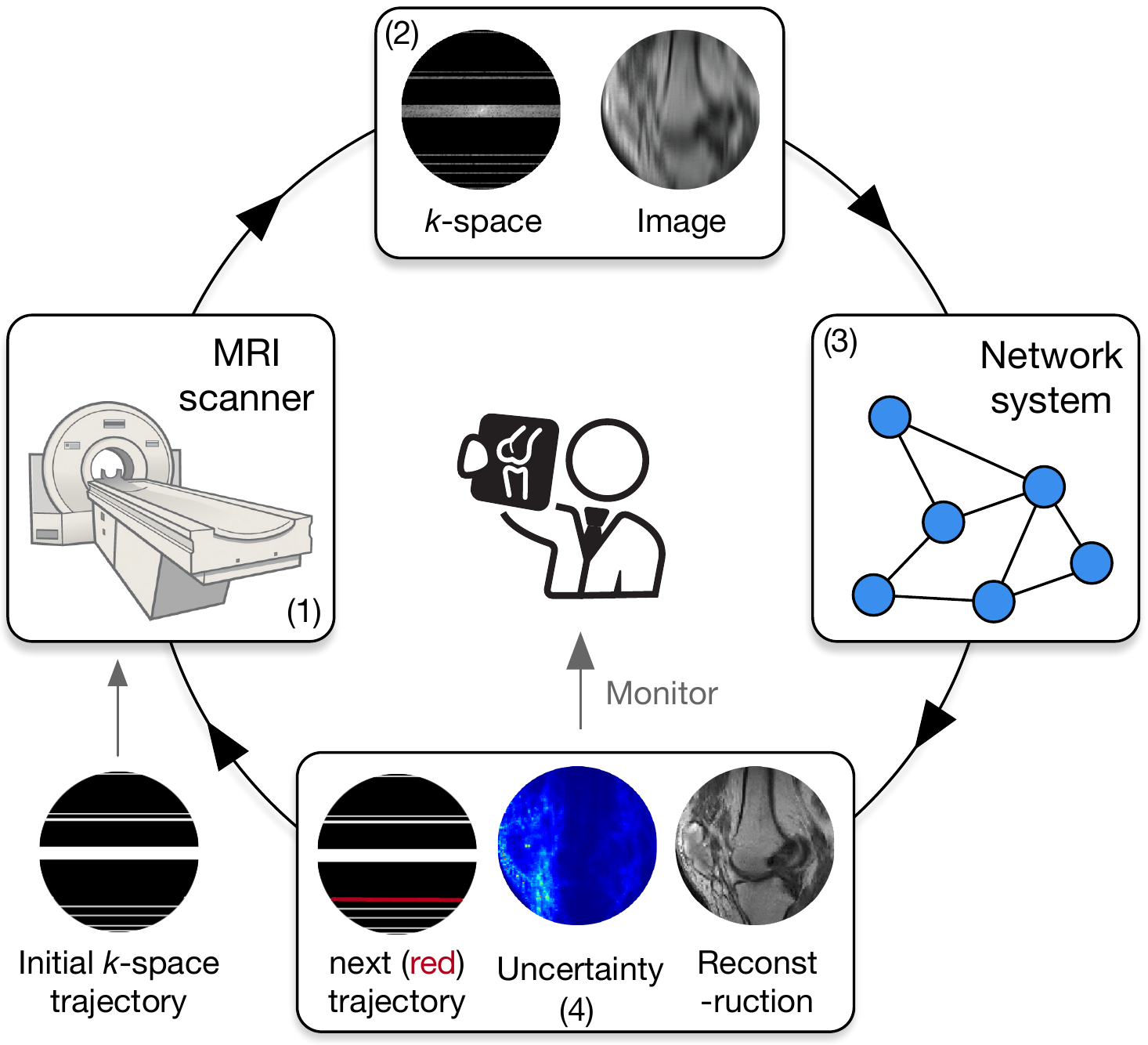}
    \vspace{-.2cm}
    \caption{Overview of our proposed pipeline. A MRI scanner (1) acquires measurements given an initial trajectory. The \emph{zero-filled} image reconstruction (2) is fed into our system (3), which outputs a reconstruction, an uncertainty map and the next suggested measurement (in red) to scan (4). These steps are repeated until the stopping criteria is met.} \label{fig:intro}  \vspace{-.2cm}
\end{figure}

Reducing the number of \textit{k}-space measurements is a standard way of speeding up the examination time. However, the images resulting from basic reconstructions from the undersampled \emph{k}-space often exhibit blur or aliasing effects \cite{moratal2008k}, making them unsuitable for clinical use. Hence, the goal of MRI reconstruction systems is to reduce the previously mentioned artifacts and recover high fidelity images.

\begin{figure}[t]
    \centering
    \includegraphics[width=0.47\textwidth]{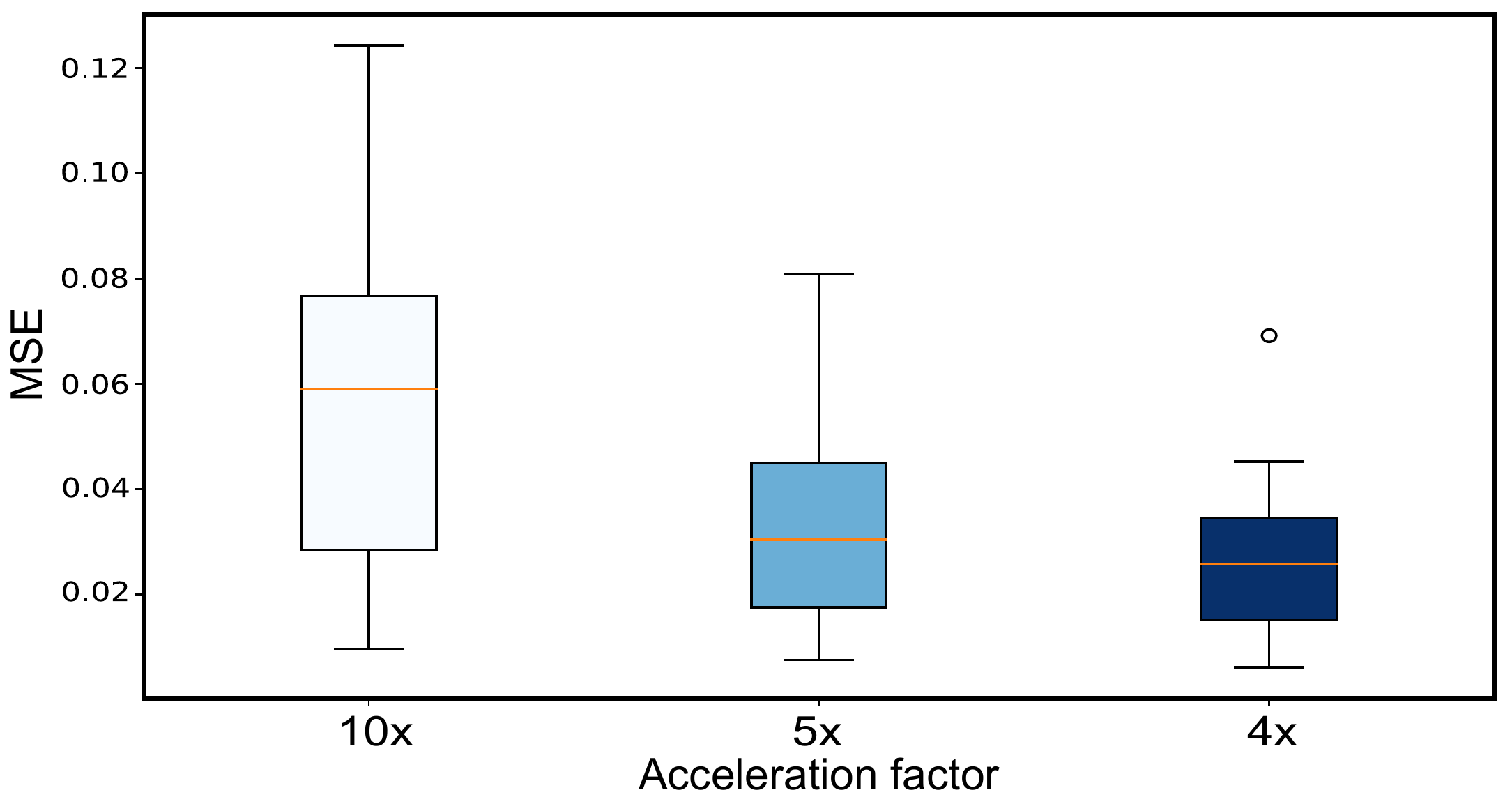}
    \vspace{-.3cm}
    \caption{Box plots representing the variance of the reconstruction mean squared errors (MSE) for different acceleration factors. To obtain the plots, we apply random \textit{k}-space trajectories with different acceleration factors to a set of images and feed them to a reconstruction network.} \label{fig:var}  \vspace{-.4cm}
\end{figure}

Deep learning has recently shown great promise in MRI reconstruction with convolutional neural networks (CNNs) \cite{hyun2018deep, schlemper2018deep, zhu2018image, Han18}. Most of these methods are designed to work with a fixed set of measurements defining a sampling trajectory\footnote{Throughout the paper, we use horizontal Cartesian acquisition trajectory, where \emph{k}-space is acquired row-by-row and we use \emph{measurement} to refer to a whole row of the Cartesian trajectory.}. We argue that this sampling trajectory should be adapted on the fly, depending on the difficulty of the reconstruction. Figure \ref{fig:var} depicts box plots obtained by applying a reconstruction network to a large dataset for three acceleration factors, namely: $10\times$, $5\times$ and $4\times$. As shown in the figure, the $10\times$ plot exhibits the highest variance. As we introduce more measurements (by reducing the acceleration factor), the error variance decreases, highlighting the existing trade-off between acquisition speedup and reconstruction error variance when fixing the sampling trajectory. A natural way to overcome this trade-off is to define data driven sampling trajectories, via \emph{active acquisition}\footnote{Note that, in active acquisition, the sampling trajectory would not only determine the number of measurements but also their sampling order.} that adapt to reconstruction difficulty by selecting sequentially which parts of \emph{k}-space to measure next.

Partial measurements naturally induce reconstruction \emph{uncertainty}, as they might be consistent with multiple, equally plausible high fidelity reconstructions, which may or may not correspond to the reconstruction from fully observed \textit{k}-space. In practice, these reconstructions could eventually mislead radiologists. Therefore, the ability to quantify and display the pixel-wise reconstruction uncertainty is of paramount relevance. On one hand, this pixel-wise uncertainty could allow radiologists to gain additional insight on the quality of the reconstruction and potentially yield a better diagnosis outcome. On the other hand, the \emph{reduction in uncertainty} via \emph{additional measurements} could be used as a signal to guide active acquisition.

In this paper, we propose a system for MRI reconstruction that, at inference time, actively acquires \emph{k}-space measurements and iteratively refines the prediction with the goal of reducing the error and, thus, the final uncertainty (see Figure \ref{fig:intro}). To do so, we introduce a novel \emph{evaluator network} to rate the quality gain in reconstruction of each \textit{k}-space measurement. This evaluator is trained jointly with a \emph{reconstruction network}, which outputs a high fidelity MRI reconstruction together with a pixel-wise uncertainty estimate. We explore a variety of architectural designs for the reconstruction network and present a residual-based model that exploits the underlying characteristics of MRI reconstruction. We extensively evaluate our method on a large scale knee MRI DICOM dataset and on ImageNet \cite{imagenet_cvpr09}. Our results show that (1) our evaluator consistently outperforms standard \emph{k}-space active acquisition heuristics on both datasets; (2) our reconstruction network improves upon common pixel-wise prediction networks and; (3) the uncertainty predictions correlate with the  reconstruction errors and, thus, can be used to trigger the halt signal to stop the active acquisition process.

To summarize, the contributions of the paper are the following:
\begin{compactitem}
\item We introduce a reconstruction network design, which outputs both image reconstruction and uncertainty predictions, and is trained to jointly optimize for both. 
\item We introduce a novel evaluator network to perform active acquisition, which has the ability to recommend  \emph{k}-space trajectories for MRI scanners and reduce the uncertainty efficiently.
\item We show through extensive evaluation the superior performance of the proposed approach, highlighting its practical value and paving the road towards improved practically applicable systems for accelerating MRI.
\end{compactitem}

\section{Related Work}
\vspace{-.1cm}
\textbf{MRI reconstruction.} There is a vast literature tackling the problem of undersampled MRI reconstruction. State-of-the-art solutions include both signal processing techniques (e.g. Compressed Sensing (CS)) as well as machine learning ones. On one hand, CS-based MRI reconstruction has been widely studied in the literature \cite{lustig2008compressed,otazo2010combination,lustig2007sparse,quan2017compressed,tygert2018compressed}. These approaches usually result in over-smoothed reconstructions, which involve a time consuming optimization process, limiting their practical scalability. On other hand, deep learning based approaches have been introduced as a promising alternative to MRI reconstruction \cite{xu2014deep, schlemper2018deep, lonning2018recurrent, hyun2018deep, ronneberger2015u}. In \cite{schlemper2018deep}, a cascaded CNN with a consistency layer is presented to ensure measurement fidelity in dynamic cardiac MRI reconstruction. In \cite{hyun2018deep}, a Unet architecture \cite{ronneberger2015u} is used to reconstruct brain images, while \cite{lonning2018recurrent} proposes a recurrent inference machine for image reconstruction. Moreover, following recent trends, architectures involving image
refinement mechanisms seem to be gaining increasing attention \cite{schlemper2018deep,seitzer2018adversarial,lonning2018recurrent}. Although all previously-mentioned approaches are able to improve the reconstruction error, the human perception of the results is still not compelling. Therefore, recent works have also focused on exploring different training objectives such as adversarial losses \cite{yang2018dagan,goodfellow2014generative,isola2017image} to enhance the perceptual reconstruction quality \cite{seitzer2018adversarial,zhang2018multi}.

\textbf{Uncertainty.} Significant effort has been devoted in the computer vision literature to provide uncertainty estimates \cite{kendall2017uncertainties} of predictions. There are two possible sources of uncertainty \cite{Kiureghian}: 1) model uncertainty due to an imperfect model (epistemic uncertainty) and 2) data uncertainty due to imperfect measurements (aleatoric uncertainty). While model uncertainty can be decreased with better models, %
data uncertainty vanishes only with the observation of all variables with infinite precision. In medical imaging, uncertainty is often used to display probable errors \cite{ching2018opportunities} and has been mainly studied in the context of image segmentation \cite{devries2018leveraging,leibig2017leveraging}. Segmentation errors (i.e. wrong label predictions) are often easier to detect by domain experts than reconstruction errors (i.e. shift of pixel values), which could potentially mislead diagnosis. Therefore, the study of uncertainty is crucial in the context of MRI reconstruction. In this paper, we focus on \emph{data uncertainty}, which is caused by the partially observed \emph{k}-space. This uncertainty can be captured by proper model parametrization, e.g. in regression tasks a Gaussian observation model is often assumed \cite{kendall2017uncertainties,kendall2017multi}; this assumption can be relaxed to allow the use of arbitrary observation models as explained in \cite{gurevich2017learning}. 

\textbf{Active acquisition.} %
Previous research on optimizing \textit{k}-space measurement trajectories from the MRI community include
CS-based techniques \cite{seeger2010optimization,ravishankar2011adaptive,zhang2014energy,gozcu2018learning}, SVD basis techniques \cite{zientara1994dynamically,panych1996implementation,zientara1998applicability}, and region-of-interest techniques \cite{yoo1999real}. %
It is important to note that all these approaches work with fixed trajectories at inference time. By contrast, \cite{levine2018fly} proposed an on-the-fly eigenvalue based approach that adapts to encoding physics specific to the object. However, contrary to our approach, it requires solving an optimization problem at inference time. Moreover, since we train all the components of our pipeline jointly, our adaptive acquisition incorporates information on the image physics, the object being imaged, and the reconstruction process to select the next measurement.

\section{Background and notation} 
\vspace{-.1cm}
Let $\mathbf{y} \in \mathbb{C}^{N{\times}N}$ be a complex-valued matrix representing the fully sampled \textit{k}-space. Neglecting effects such as magnetic field inhomogeneity and spin relaxation, the image can be estimated from the \textit{k}-space data by applying a 2D Inverse Fast Fourier Transform (IFFT) $\mathbf{x} = \mathcal{F}^{-1}(\mathbf{y})$, where $\mathbf{x} \in \mathbb{C}^{N{\times}N}$ is the image and $\mathcal{F}^{-1}$ is the IFFT operation. We denote the binary sampling mask defining the \textit{k}-space Cartesian acquisition trajectory as $\mathbf{S}$ \cite{zhu2018image}. The acquired measurements are referred to as \emph{observed} whereas the masked measurements are referred to as \emph{unobserved}. We define the undersampled, partially observed \textit{k}-space as $\mathbf{\hat{y}} = \mathbf{S \odot y}$, where $\odot$ denotes element-wise multiplication. Thus, the basic \emph{zero-filled} image reconstruction is obtained as $\mathbf{\hat{x}} = \mathcal{F}^{-1}(\mathbf{\hat{y}})$. Analogously, we can go from the reconstructed image to the \textit{k}-space measurements $\mathbf{\hat{y}} = \mathcal{F}(\mathbf{\hat{x}})$, where $\mathcal{F}$ is the Fast Fourier Transform (FFT).

It is worth noting that MRI images $\mathbf{x} = \mathcal{F}^{-1}(\mathbf{y})$ are complex-valued matrices. However, most Picture Archiving and Communication Systems in hospitals do not store raw \emph{k}-space measurements, but instead store the magnitude image $abs(\mathbf{x}) \in \mathbb{R}$ in the DICOM format. Therefore, we simulate \textit{k}-space measurements by applying the FFT to the magnitude image $\mathbf{y} = \mathcal{F}(abs(\mathbf{x}))$. We do not differentiate the notation of an image in $\mathbb{R}$ or $\mathbb{C}$ hereinafter.

We make use of one of the numerous properties of FFT\footnote{See Chapter $3.4$ of \cite{szeliski2011computer} for the full list.}, namely Parseval's Theorem \cite{rippel2015spectral}. %
It implies that the $l_2$-distance between two images $\mathbf{x}^{(1)}, \mathbf{x}^{(2)}$ is equivalent to the $l_2$-distance between their representation in the frequency domain, i.e. $||\mathcal{F}(\mathbf{x}^{(1)}) - \mathcal{F}(\mathbf{x}^{(2)})||^2_2 =  ||\mathbf{x}^{(1)} - \mathbf{x}^{(2)}||_2^2$.

\label{sec:method}
\begin{figure}[t]
    \centering
    \includegraphics[width=0.499\textwidth]{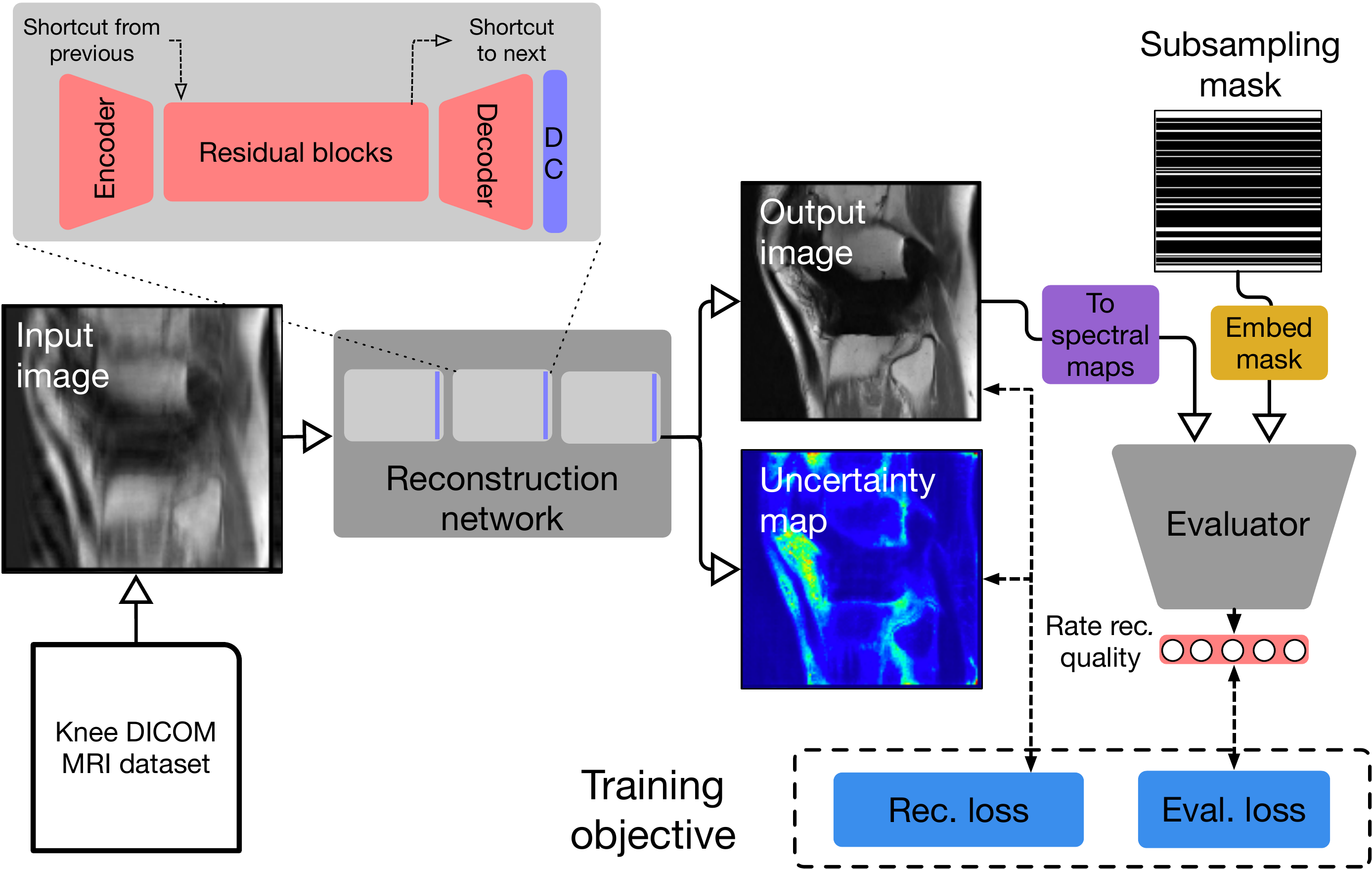}
    \vspace{-.5cm}
    \caption{The training pipeline of the proposed method.} \vspace{-.5cm} \label{fig:arch} 
\end{figure}

\section{Method}
\vspace{-.1cm}
Figure \ref{fig:arch} illustrates our approach. The framework is composed of (1) a \emph{reconstruction network} and (2) an \emph{evaluator}. The goal of the reconstruction network is to produce high fidelity reconstructions from undersampled \emph{k}-space measurements. The network takes a basic \emph{zero-filled} image reconstruction as input and outputs an improved image reconstruction together with its uncertainty estimates. The goal of the evaluator network is to rate each corresponding \emph{k}-space row of a reconstructed image, where the score should indicate how much it resembles true measurements. The rating score guides the measurement selection criterion: the lowest rated measurement should be acquired first.

\subsection{Reconstruction network}
\vspace{-.1cm}
Our reconstruction network has a cascaded backbone composed of residual networks (ResNets) \cite{he2016deep}, more precisely fully convolutional ResNets (FC-ResNets) \cite{DrozdzalVCKP16, Casanova18} followed by \emph{data consistency} (DC) layers \cite{schlemper2018deep}. %

The DC layer \cite{schlemper2018deep}\footnote{We use the noiseless version of DC, which makes $\mathcal{F}(\mathbf{\hat{x}})$ fully preserved in the output, with a hard copy. See \cite{schlemper2018deep} for more details.} builds direct shortcut connections from the input of the network $\mathbf{\hat x}$ to its output $f(\mathbf{\hat{x}})$ to enforce the preservation of the observed information while estimating the reconstruction. The DC layer operates in \emph{k}-space, and the reconstruction can be formally defined as:
\vspace{-.1cm}
\begin{equation} 
  \mathbf{r} = \mathrm{DC}(\mathbf{\hat x}, \mathbf{S}) = \mathcal{F}^{-1}((1-\mathbf{S}) \odot \mathcal{F}(f(\mathbf{\hat x})) +  \mathbf{S} \odot \mathcal{F}(\mathbf{\hat x})). \vspace{-.1cm}
\end{equation}

The rationale behind choosing FC-ResNet followed by DC layers as building block of our cascaded network is to learn the residual $f(\mathbf{\hat{x}}) = \mathbf{r} - \mathbf{\hat{x}}$. Thus, $f(\mathbf{\hat{x}})$ estimates the image representing the unobserved part of $\mathcal{F}(\mathbf{x})$, complementing $\mathcal{F}(\mathbf{\hat{x}})$. The rationale behind cascading the previously described building blocks is to provide intermediate deep supervision \cite{LeeXGZT15}. 

Overall, the proposed cascaded FC-ResNet (denoted c-ResNet) concatenates three identical tiny encoder-decoder networks, interleaved with DC layers. Note that this network is reminiscent of the 3D cascaded CNN proposed in \cite{schlemper2018deep} with minor design changes and endowed with deep supervision. To enhance the information flow between FC-ResNet modules, we add a shortcut to link residual blocks between adjacent modules (Figure \ref{fig:arch}). Hence, each module can re-use the representations of its predecessor and enhance the representations with further network capacity (see the supplementary materials for details). 

\subsection{Uncertainty estimates}
\vspace{-.1cm}
FC-ResNet modules described in the previous section are trained to also output pixel-wise uncertainty estimates $u(\mathbf{\hat x})$, which we will use to trigger the halt signal to stop the active acquisition process. The additional benefit of having uncertainty estimates is that they highlight regions of the image that are likely to contain large reconstruction errors. %
Similarly to \cite{gurevich2017learning,kendall2017uncertainties}, we model the uncertainty about the value of a pixel as a Gaussian centered at reconstruction mean $\mathbf{r}$ and with variance $u(\mathbf{\hat{x}})$, i.e. $\mathcal{N}(\mathbf{r}, \mathrm{diag}(u(\mathbf{\hat{x}})))$. We train our reconstruction network to maximize the average conditional log-likelihood, which amounts to minimizing:
\begin{equation}
\mathcal{L}_R(\mathbf{\hat x}, \mathbf{r}, \mathbf{x}) = \frac{1}{N^2} \sum_{i=1}^{N^2} \frac{|\mathbf{r}_i - \mathbf{x}_i|^2}{2 u(\mathbf{\hat{x}})_i}  + \frac{1}{2} \log(2\pi u(\mathbf{\hat{x}})_i), \vspace{-.05cm}
\end{equation}
where $\mathbf{x}$ is the "ground-truth" target image, $\mathbf{\hat{x}}$ is a \emph{zero-filled} reconstruction given as input to the network, $\mathbf{r}$ is the reconstruction it outputs, and $N^2$ is the number of pixels.

\subsection{Evaluator network}
\label{ssec:evaluator}
\begin{figure}[t]
    \centering
    \includegraphics[width=0.49\textwidth]{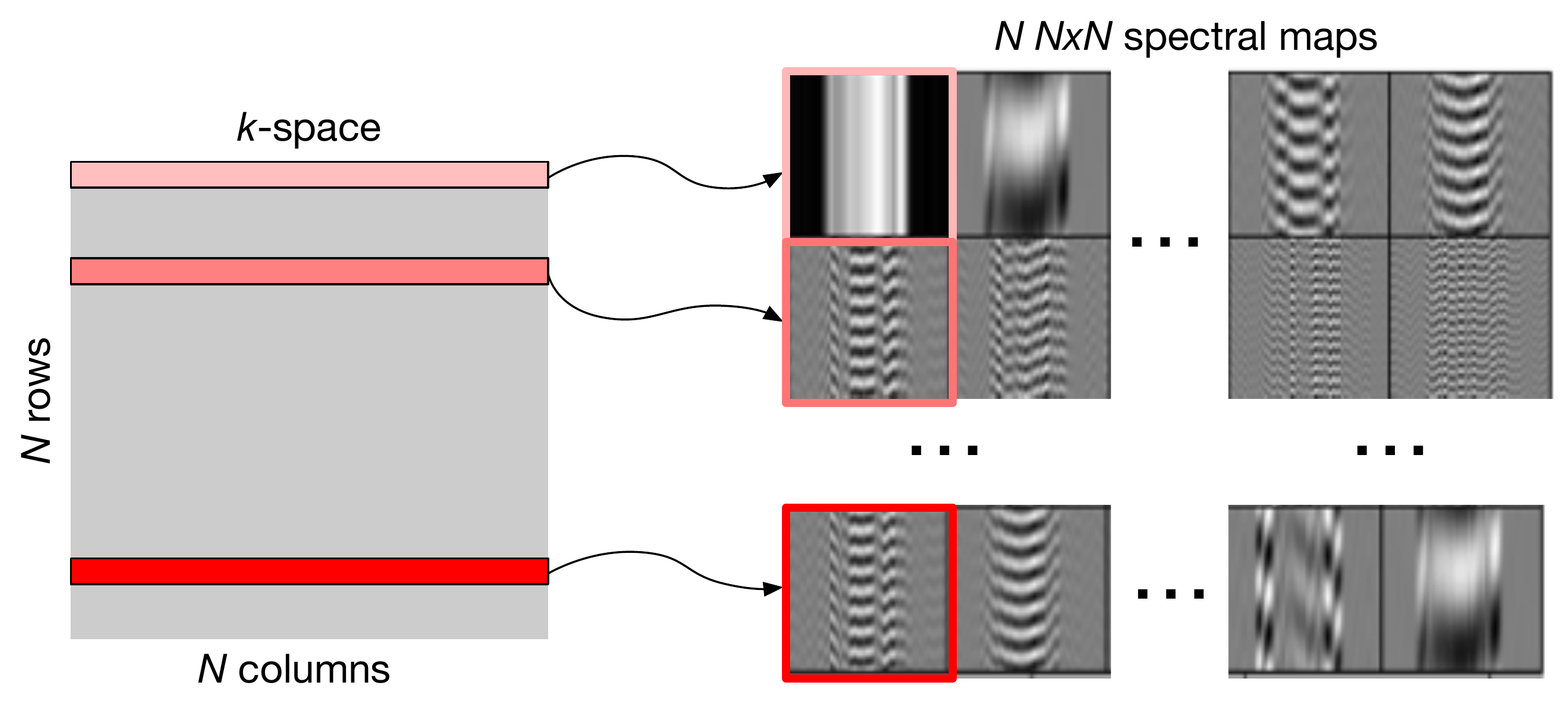}
    \vspace{-.7cm}
    \caption{Image decomposition into $N$ spectral maps. \vspace{-.4cm}} \label{fig:sectral_map} 
\end{figure}

\begin{figure*}[t] %
    \begin{minipage}[b]{0.299\linewidth}
        \begin{tabularx}{.99\textwidth}{c|cc}
            \hline  
            Method   & MSE & SSIM \\ \hline
            pix2pix    &  0.100  & 0.61  \\ 
            FC-DenseNet &  0.072  & 0.70 \\ 
            Unet    &  0.065   &  0.72 \\ 
            ResNet &   0.055  & 0.75  \\ \hline
            Ours (c-ResNet) &  \textbf{0.050}   & \textbf{0.77} \\
            Ours    &  0.052 & 0.76 \\ 
            \hline
        \end{tabularx}
        \vspace{1.0cm}
        \captionof{table}{MSE /SSIM at kMA = 21\%.}\label{tab:rec_mse}
    \end{minipage}
    \begin{minipage}[b]{0.68\linewidth}
        \includegraphics[width=0.99\textwidth,height=0.4\textwidth]{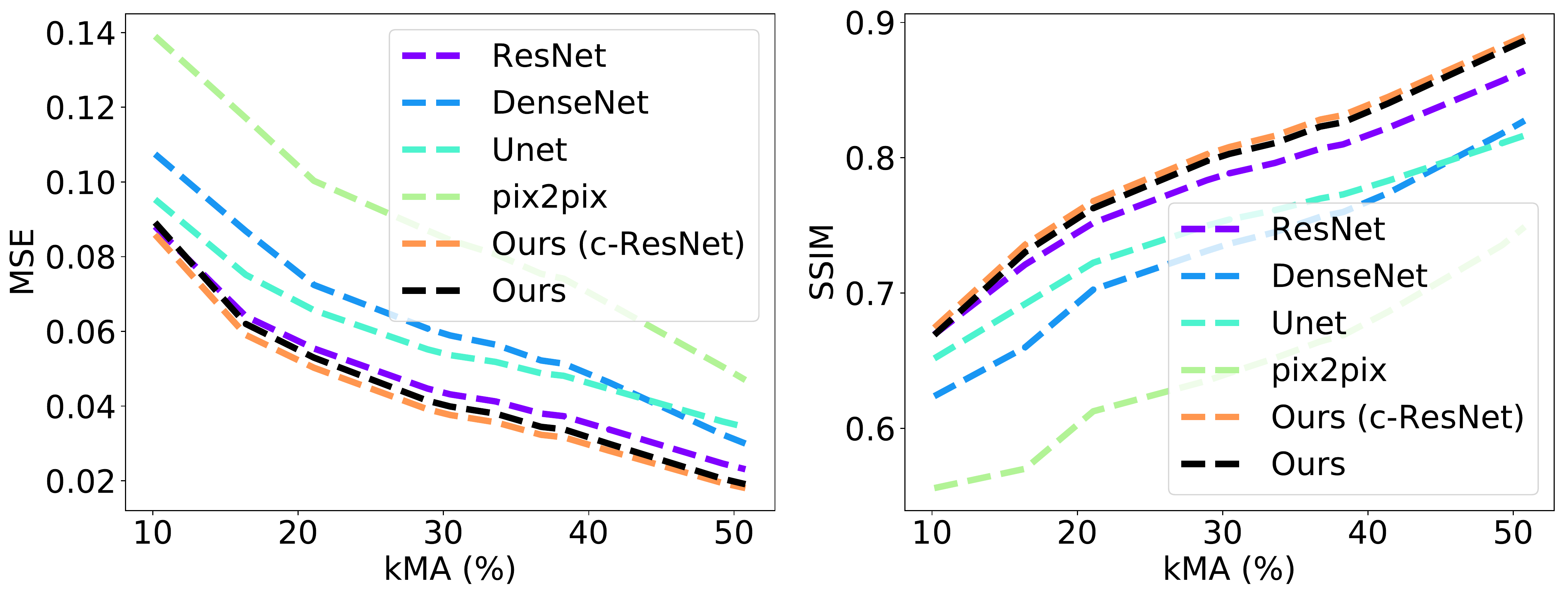} 
        \vspace{-0.2cm}
        \caption{Plots depicting MSE and SSIM for different kMA values.} \label{fig:rec_mse}
    \end{minipage} %
    \vspace{-.2cm}
\end{figure*}

\begin{figure*}[t]
    \centering
    \includegraphics[width=0.99\textwidth]{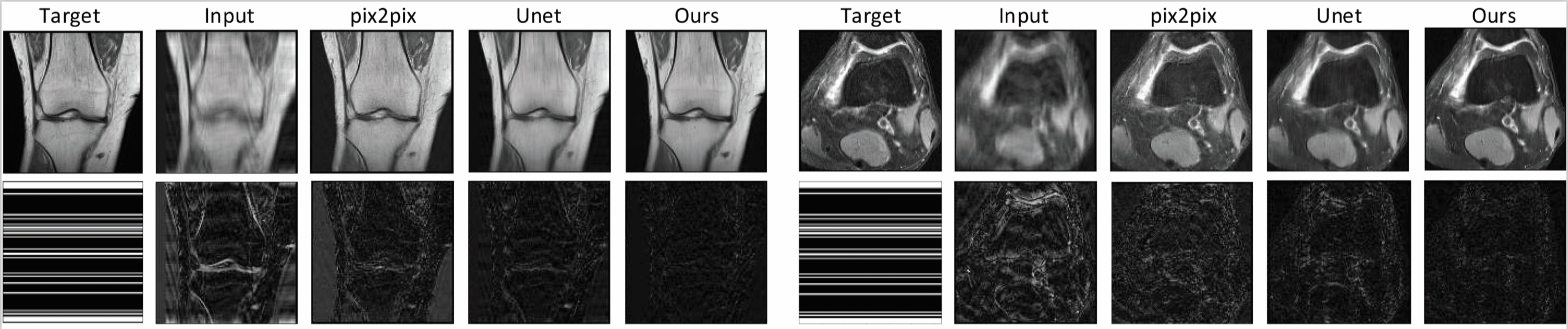}
    \vspace{-.3cm}
    \caption{Qualitative comparison of different reconstruction networks, including reconstruction results and error maps (normalized for improved visualization). The binary image below target is the sampling trajectory with $25\%$ kMA.} \vspace{-.3cm} \label{fig:rec_comp} 
\end{figure*}

The role of the evaluator network $e$ is to tell whether a given k-space row is likely to be a true $k$-space measurement or to come from a reconstruction. When training the reconstruction network, we will be using the evaluator as additional regularization to encourage the reconstructed image to have phantasized $k$-space rows that look as if they came from the distribution of true measured rows. To be proficient in this task, the evaluator has to be able to capture small structural differences in images that define the distribution of the true, observed measurements. 
In our design, we leverage the idea of adversarial learning \cite{goodfellow2014generative,RadfordMC15}, and train a discriminator-like evaluator to score the measurements and meanwhile encourage the reconstruction network to produce results that match the true measurement distribution.

The first step of the evaluator decomposes the output image reconstruction $\mathbf{r} \in \mathbb{C}^{N{\times}N}$ into $N$ \emph{spectral maps}, each one corresponding to a \emph{single} $k$-space row. 
To obtain these spectral maps, we first transform $\mathbf{r}$ into the \textit{k}-space representation $\mathbf{y}=\mathcal{F}(\mathbf{r})$. Then, we mask out all the \textit{k}-space rows except of the $i$-th one using a binary mask $\mathbf{\hat{S}}^{(i)}$. The $i$-th spectral map of a reconstruction output is obtained as $\mathbf{M(\mathbf{r})}^{(i)} = \mathcal{F}^{-1}(\mathbf{\hat{S}}^{(i)}\odot\mathcal{F}(\mathbf{r}))$. %
Analogously $\mathbf{M(\mathbf{x})}^{(i)}$ denotes the $i$-th spectral map of the ground truth reconstruction~\footnote{Note that using the linearity of the Fourier transform, one could write: $\mathbf{r} = \sum_{i}^{N} \mathbf{M(\mathbf{r})}^{(i)}$.}. This process is depicted in Figure \ref{fig:sectral_map}. Moreover, it embeds the acquisition trajectory $\mathbf{S}$ into a $6$D vector. Finally, both the spectral maps and the trajectory embedding as a 3D tensor are fed to a CNN, whose full architectural details are provided in the supplementary material.

We train the evaluator so that it assigns a high value to spectral maps that correspond to actually observed rows of the \textit{k}-space and a low value to the unobserved ones. The simplest approach would be to train a discriminator to distinguish between observed and unobserved rows. However, we found that such strategy does not work well: the evaluator tends to output polarized scores (close to $0$ or $1$), making it hard to use to rank unobserved measurements. 
Instead, we decompose both the ground truth image $\mathbf{x}$ and the reconstruction output $\mathbf{r}$ into spectral maps and train the evaluator network $e(\mathbf{r}, \mathbf{S})$ to fit target scores given by the following kernel:
\vspace{-.2cm}
\begin{equation} 
     t(\mathbf{r}, \mathbf{x})_i = \exp(- \gamma ||\mathbf{M(\mathbf{r})}^{(i)} - \mathbf{M(\mathbf{x})}^{(i)}||_2^2), \vspace{-.2cm}
\end{equation}
where $\gamma$ is a scalar hyper-parameter. 
Specifically $e$ is trained to minimize the following objective:
\vspace{-.2cm}
\begin{equation} 
\label{eq:eval}
     \mathcal{L}_E^E(\mathbf{r}, \mathbf{x}, \mathbf{S}) = \sum_{i}^{N} |e(\mathbf{r}, \mathbf{S})_i - t(\mathbf{r}, \mathbf{x})_i|^2, \vspace{-.2cm}
\end{equation}
where $e(\mathbf{r}, \mathbf{S})_i$ is the score of measurement $i$.
Note that $t(\mathbf{r}, \mathbf{x})_i$ is close to $1$ when $\mathbf{M(\mathbf{r})}_i$ is similar to $\mathbf{M(\mathbf{x})}_i$ and is close to $0$ otherwise\footnote{Thus, $t_i$ can be viewed as an energy function \cite{zhao2016energy} we expect to minimize by updating the parameters of the reconstruction network.}. Note that the DC layer always ensures that $\mathbf{M(\mathbf{r})}_i$ is equal to $\mathbf{M(\mathbf{x})}_i$ for the observed rows of the \textit{k}-space. Hence $t_i \equiv 1$ for observed measurements.

\subsection{Joint adversarial training}
\vspace{-.1cm}

Following the principle of adversarial training, the evaluator network is used to update the reconstruction network using the following objective:
\vspace{-.3cm}
\begin{equation} 
     \mathcal{L}_E^R(\mathbf{r}, \mathbf{S}) = \sum_{i}^{N} |e(\mathbf{r}, \mathbf{S})_i - 1|^2, \vspace{-.3cm}
\end{equation}
which encourages the reconstruction network to produce reconstructions that can obtain high evaluator scores $e(\mathbf{r}, \mathbf{S})$. 
Overall, the reconstruction network is trained with the following objective:
\vspace{-.2cm}
\begin{equation} 
     \mathcal{L}(\mathbf{R}, \mathbf{x}, \mathbf{S}) = \frac{1}{K}\sum_{k=1}^{K} \mathcal{L}_R^k(\mathbf{r}^{k-1}, \mathbf{r}^k, \mathbf{x}) + \beta\mathcal{L}_E^R(\mathbf{r}^{K}, \mathbf{S}),
\label{eq:full_objective} \vspace{-.2cm}
\end{equation}
where $\mathbf{R} = [\mathbf{r}^0, ..., \mathbf{r}^K]$, $\mathbf{r}^0 = \mathbf{\hat x}$, $\mathbf{r}^k$ for $k \geq 1$ is the output of the $k$-th cascading block, $\beta$ is a hyper-parameter controlling the influence of the evaluator loss on the global objective and $K$ is the number of cascaded FC-ResNets in the reconstruction network.

We train the full model end-to-end, by alternating the reconstruction and evaluator networks' updates as in the standard adversarial training fashion \cite{goodfellow2014generative}. We use the Adam solver ($\beta_1=0.5$, $\beta_2=0.999$) \cite{kingma2014adam} with an initial learning rate of $0.0006$ for $50$ epochs. The learning rate is then linearly decreased per epoch for another 50 epochs, until it reaches 0. For all experiments, we set $\beta=0.1$, $K=3$ and $\gamma=100$. All models are trained using $6$ Tesla P100 GPUs, with a batch size of $48$ per GPU. 

\subsection{Active acquisition}
As illustrated in Figure \ref{fig:intro}, at \emph{inference time}, the evaluator scores $e(\mathbf{r}, \mathbf{S})$ are used to select the next unobserved measurement to acquire. Then, the input image is updated accordingly and the process iterates until all measurements are acquired or a stopping criteria is met, e.g. a low global uncertainty score.  

\section{Experiments}
\vspace{-.1cm}

In this section, we provide an in depth analysis of all the components of the proposed active acquisition pipeline. All experiments are conducted on a large scale Knee DICOM dataset from \cite{zbontar2018fastMRI} as well as on ImageNet \cite{deng2009imagenet}.

The Knee DICOM dataset is composed of 10k volumes. In our experiments, we use a subset of the data set and slice images from three axials at close-to-central positions of volumes, resulting in 11049 training images and 5048 test images. Among the training images, 10\% are used for validation for hyperparameter search. We report results on the test set. All images are resized to have resolution $128 \times 128$. Volumes are from different machines and they have different intensity range. We standardize each image using mean and standard deviation computed on the corresponding volume.

In order to evaluate the quality of reconstruction on a downstream classification task, we use the ImageNet dataset \cite{deng2009imagenet}. We pre-process the dataset in order to have gray scale images of $128{\times}128$ pixels. Since we can not apply any off-the-shelf RGB pre-trained classification model, we train a ResNet50 \cite{he2016deep} on the pre-processed images\footnote{We use the following implementation: \url{https://github.com/pytorch/examples/tree/master/imagenet}}. 

The training acquisition trajectory $\mathbf{S}$ is obtained following the Cartesian sampling by fixing 10 low frequency measurements in top and bottom rows and randomly sampling from the remaining ones until a desired number of measurements is obtained. In our experimental setup, the desired number of measurements is randomly chosen between 13 and 47. To evaluate the system, we characterize the acquisition trajectory $\mathbf{S}$ with the number of observed \textit{k}-space measurements w.r.t. the total number of possible measurements as $\mathrm{kMA} = \frac{\text{\# of acquired measurements}}{\text{\# of all possible measurements}}$\footnote{For DICOM data, we define the number of all possible measurements to be $N/2$ - the true degrees of freedom of our data due to the Fourier Transform's conjugate symmetry property. See supplementary material for details.}. Since, acquisition time in MRI is proportional to the number of measurements acquired, the acceleration factor is computed as $\frac{1}{\mathrm{kMA}}$. Thus, the lower kMA the higher acceleration factor (e.g. $25\%$ kMA implies a speedup of 4x). 

In the remainder of the section, we analyze the different components of our model, highlighting the obtained competitive results and its practical values. 
\begin{figure}[tb]
    \centering
    \includegraphics[width=0.4\textwidth,height=0.3\textwidth]{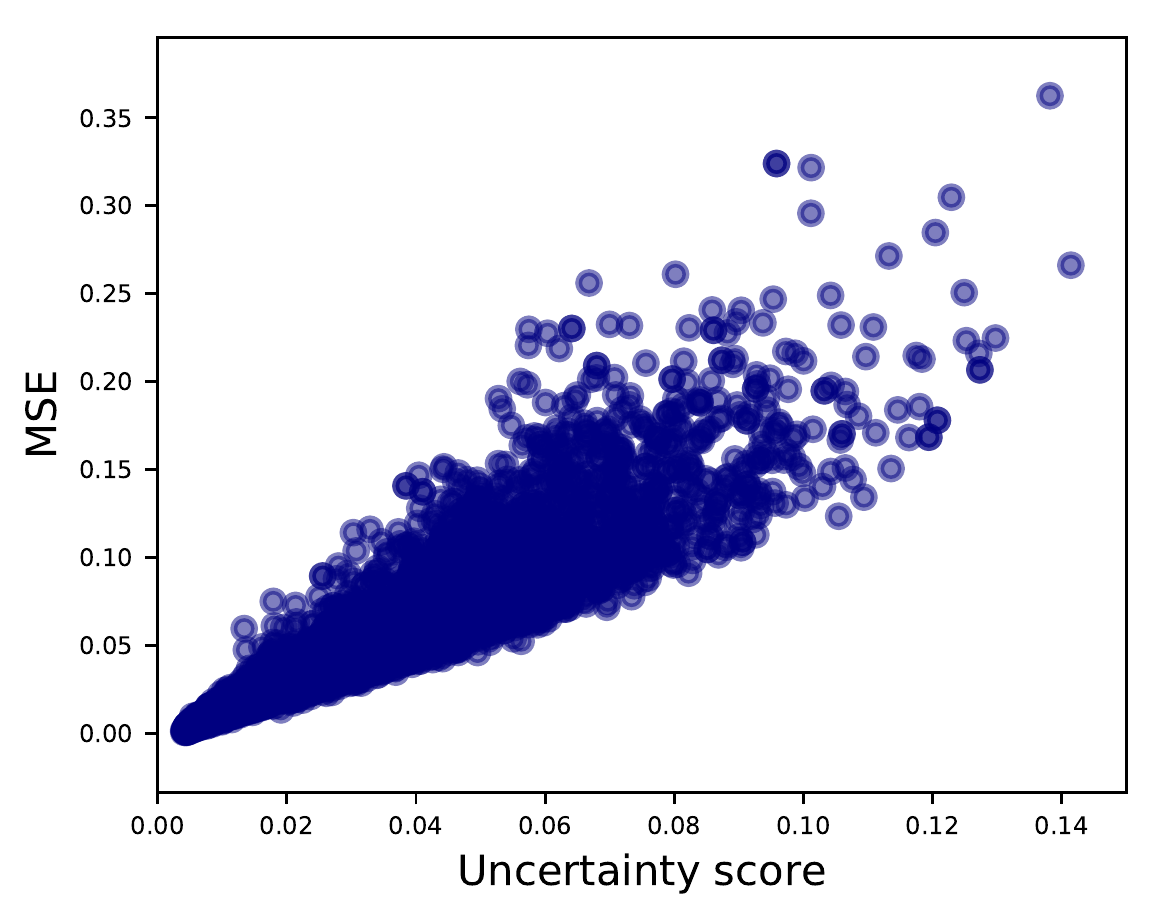}
    \vspace{-.4cm}
    \caption{Correlation plot between MSE and the mean uncertainty score, each dot represents one image.} \label{fig:cor}  \vspace{-.3cm}
\end{figure}

\begin{figure*}[t]
    \centering
    \includegraphics[width=0.999\textwidth]{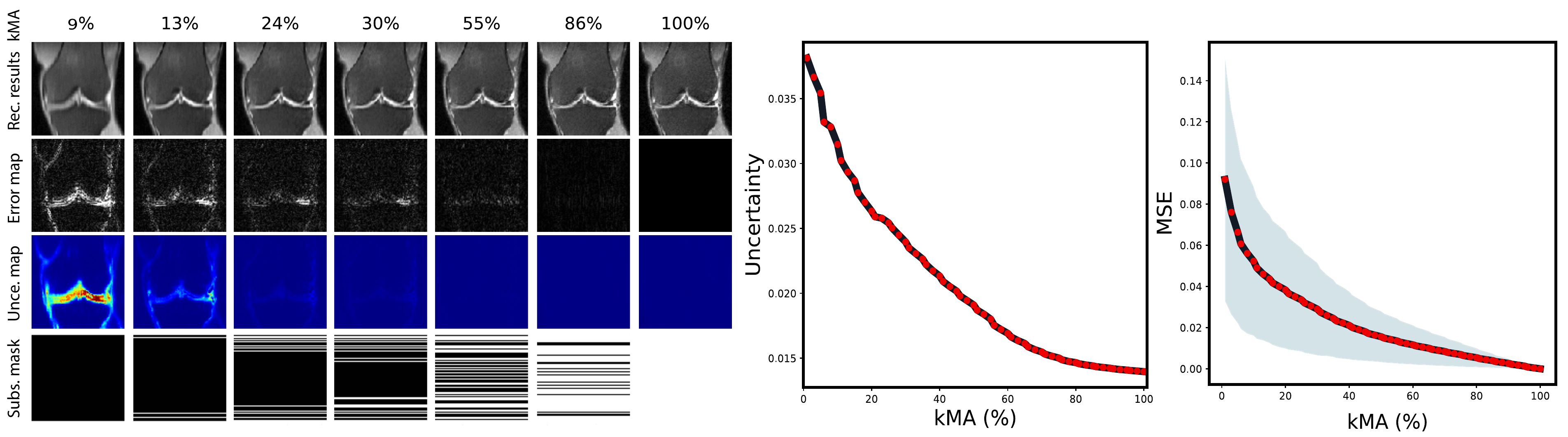}
    \vspace{-.7cm}
    \caption{Simulation of \textit{k}-space acquisition at the inference time. The left panel shows (top to bottom): reconstruction results, error maps, uncertainty maps, and sampling trajectories (in DFT coordinates). The initial mask includes $10$ low-frequency rows (in white). The plots on the right monitors both MSE and the mean uncertainty value at different kMA ratios.} \vspace{-.4cm} 
    \label{fig:simulation} 
\end{figure*}

\subsection{Comparison of reconstruction architectures}
\vspace{-.1cm}
In this subsection, we build two variants of our reconstruction architecture: (1) a vanilla c-ResNet trained by removing both the uncertainty estimates and the evaluator to minimize the mean squared error (MSE); and (2) a c-ResNet trained within the whole pipeline as described in Section \ref{sec:method}. We compare these architectures to state-of-the-art deep learning models, commonly used in the MRI literature (Unet \cite{hyun2018deep} and ResNet defined in CycleGAN \cite{zhu2017unpaired}) and in dense prediction problems (FC-DenseNet103 \cite{jegou2017one}, pix2pix \cite{isola2017image, yang2018dagan}). Note that pix2pix includes additional adversarial losses. We use MSE and Structural Similarity Index (SSIM) \cite{wang2004image} as evaluation metrics. 

For the sake of fair comparison, we add a DC layer to all models. Moreover, we found that that batch normalization (BN) \cite{ioffe2015batch} works poorly for undersampled MRI reconstruction, whereas instance normalization (IN) \cite{ba2016layer} is an important operation to improve results. Our findings are aligned with the recent work of \cite{pan2018two}, which suggests that IN learns
features that are invariant to appearance changes, while BN better preserves content related information. Therefore, we endow all models with IN instead of BN and tune them to improve performance.

Table \ref{tab:rec_mse} reports MSE and SSIM performance for all above-mentioned models at kMA = $21 \%$ ($\sim$ 5x speedup). We observe that ResNet-based architectures outperform Unet and FC-DenseNet. As shown in the table, our vanilla reconstruction network (Ours (c-ResNet)) outperforms all above-mentioned pixel-wise baselines in terms of MSE and SSIM. Our full method (Ours) also optimizes uncertainty estimates and evaluator to perform active acquisition, which hinders the direct optimization of MSE and thereby results in a slight performance drop. Similarly, the weak performance of pix2pix could be explained by its discriminator. %

Figure \ref{fig:rec_mse} depicts the MSE and SSIM performance metrics as a function of kMA. To validate the models, we create multiple validation sets by varying number of observed measurements from $10\%$ to $50\%$ kMA. All results were obtained with a single model trained on random acquisition trajectories with kMA varying from $10\%$ to $37\%$. From these experiments, we observe the same trend as reported before, namely ResNet-based architectures being better suited to perform undersampled MRI reconstruction, for all kMA values. Moreover, we can observe that all the tested models scale gracefully to unseen kMAs, namely from $38\%$ to $50\%$. Finally, we illustrate some qualitative results in Figure \ref{fig:rec_comp}.

\begin{figure}[tb!]
    \centering
    \includegraphics[width=0.42\textwidth, height=0.38\textwidth]{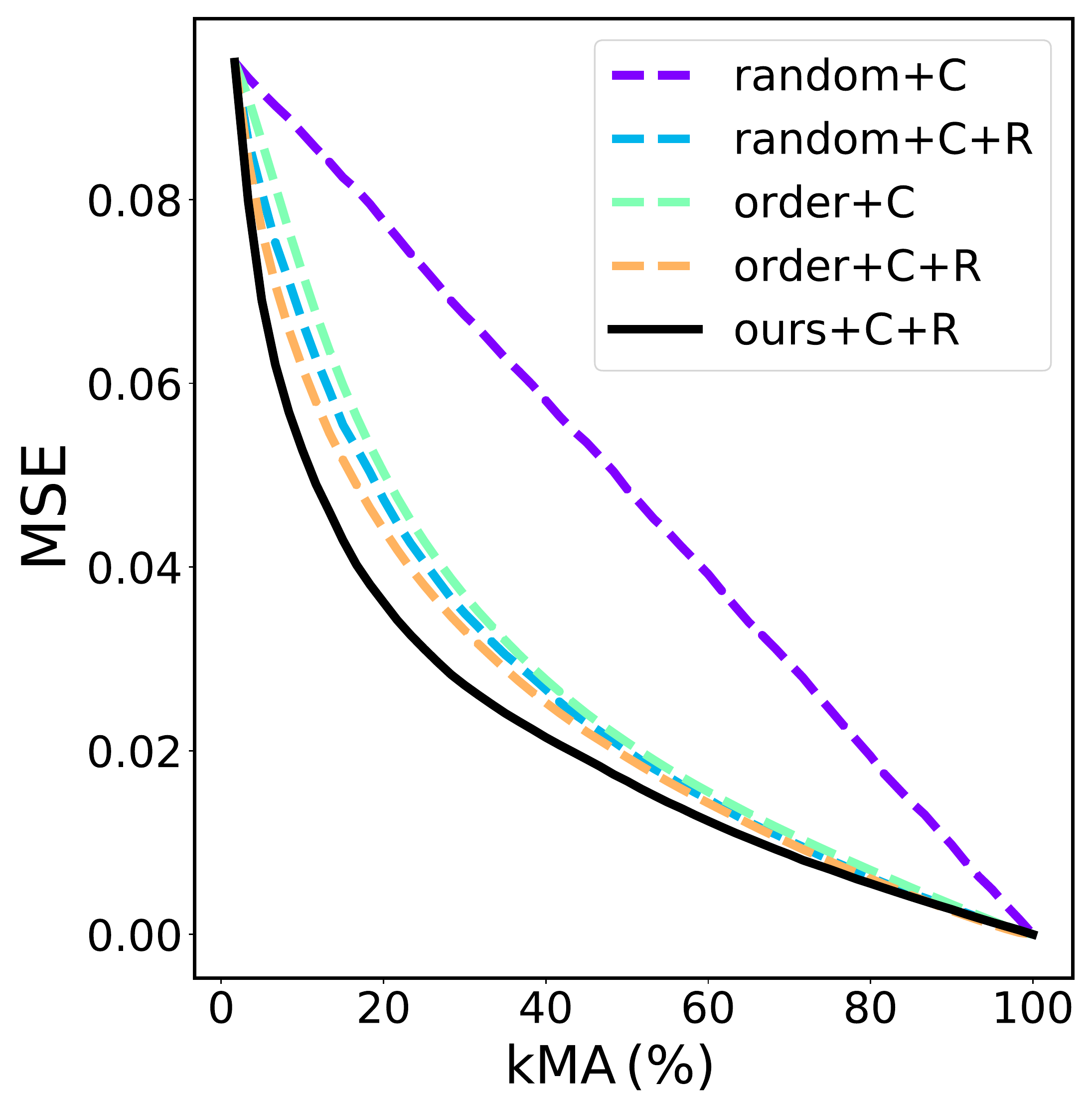}
    \vspace{-.3cm}
    \caption{Comparison of different \textit{k}-space acquisition heuristics to our model on the Knee dataset. The plot depicts MSE as a function of number of measurements.}  \vspace{-.5cm} 
    \label{fig:perf} 
\end{figure}
\begin{figure*}[]
    \centering
    \includegraphics[width=0.96\textwidth, height=0.3\textwidth]{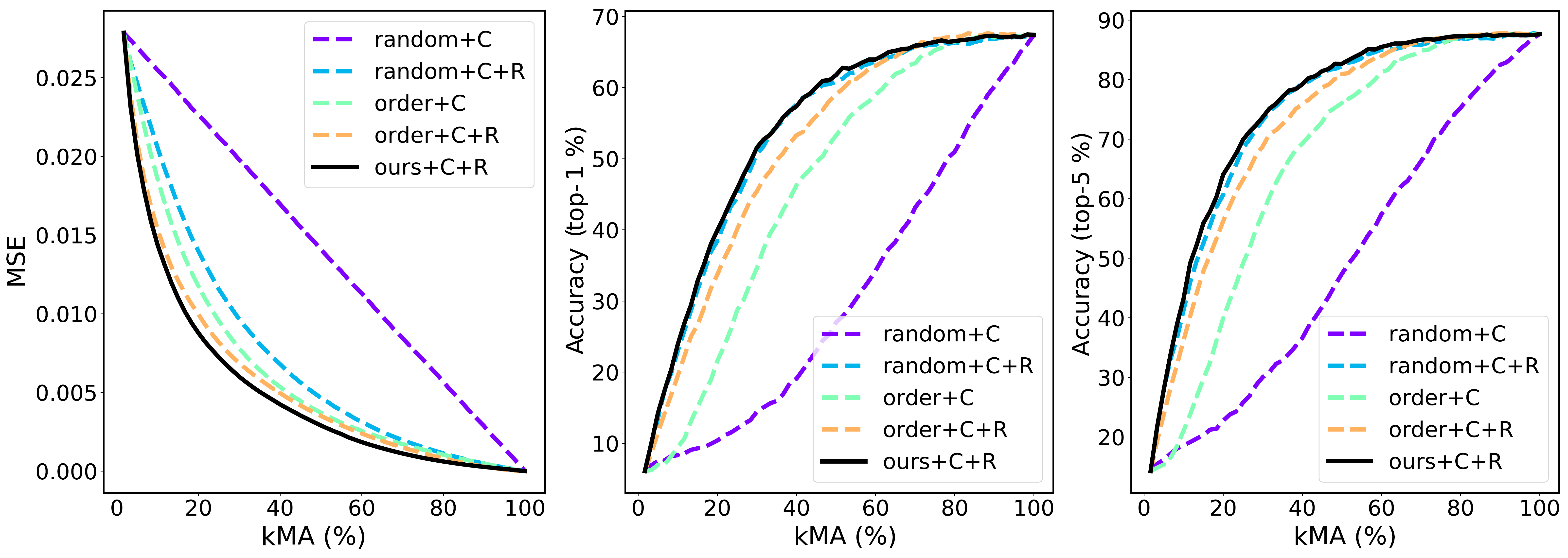}
    \vspace{-.3cm}
    \caption{Comparison of different \textit{k}-space acquisition heuristics to with our pipeline on ImageNet. The plots depict MSE and accuracy (top-$1\&5$) as a function of number of measurements. \vspace{-.7cm}} \label{fig:perf_imgnet} 
\end{figure*}
\begin{figure}[]
    \centering
    \includegraphics[width=0.4\textwidth, height=0.25\textwidth]{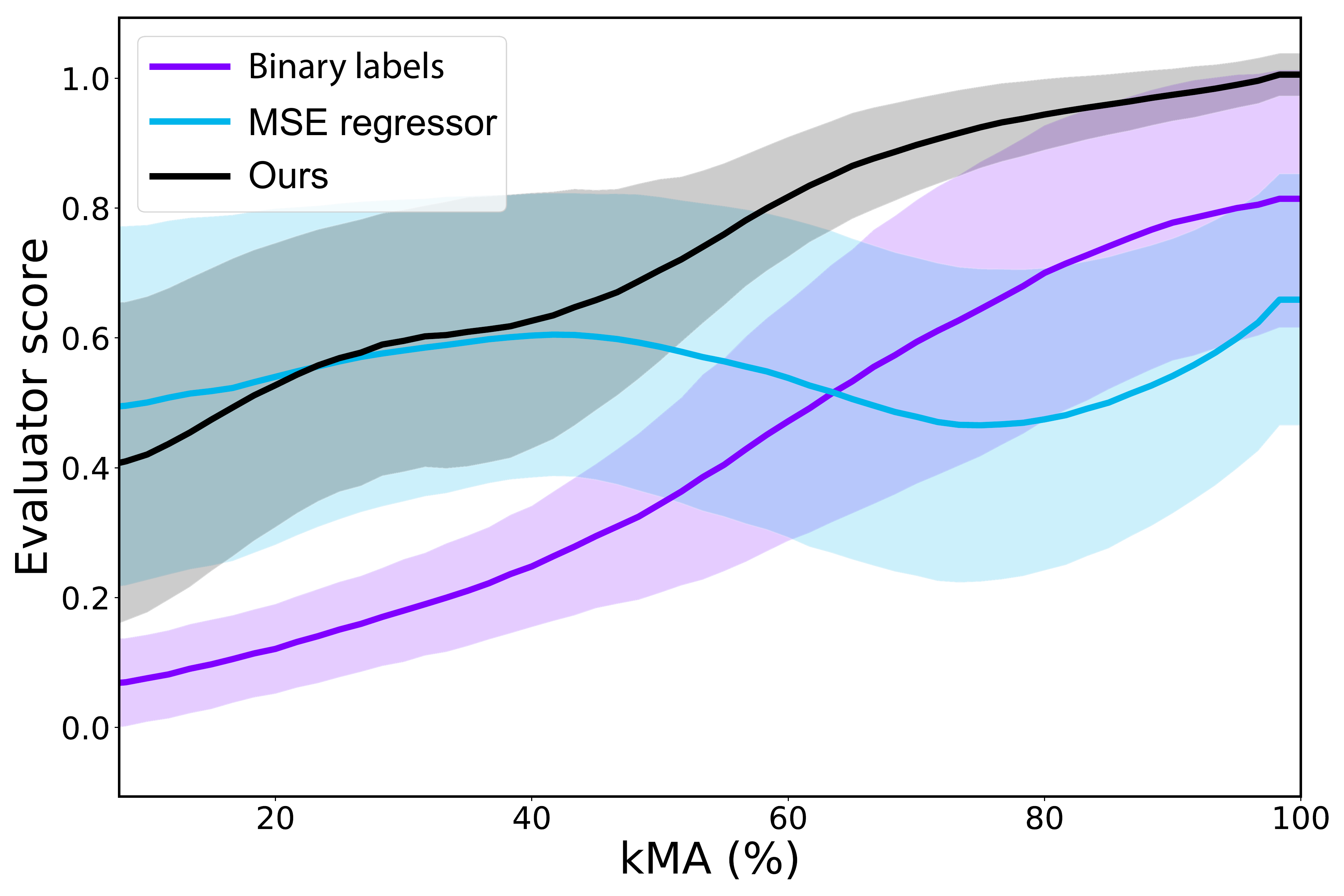}
    \vspace{-.4cm}
    \caption{Evaluator score as a function of the number of measurements. We compare our evaluator design to two baselines: MSE regressor and adversarial loss trained with binary labels. }
    \vspace{-.77cm}
    \label{fig:evaluator} 
\end{figure}

\subsection{Uncertainty analysis}
\vspace{-.1cm}
\label{ssec:uncertainty}
The goal of this subsection is to delve into the estimated uncertainty estimates and their correlation with the reconstruction errors. We select 512 test images, apply random random acquisition trajectories with kMA ranging from [10\%, 95\%], feed them to our reconstruction network and output both high fidelity reconstructions and uncertainty maps. Next, we compute the MSE between the obtained reconstructions and their corresponding ground truths as well as their mean uncertainty score. Figure \ref{fig:cor} shows the resulting correlation plot. As it can be seen, the the mean uncertainty score correlates well with the MSE.
We observe that the correlation is weaker as both MSE and uncertainty increase. These results indicate that the uncertainty estimates of our system could be useful to monitor the quality of reconstruction throughout our active acquisition process.

\subsection{\textit{k}-space active acquisition analysis}
\vspace{-.1cm}
Simulating the active acquisition process of an MRI scanner is straightforward. Given an input with a certain acquisition trajectory, we firstly obtain the reconstructed image. Then, we select the next unobserved row to acquire and measure it by copying it from the ground truth to the input image. After that, the updated input image is processed by our system. We iterate this process until the the stopping criteria is met or the \textit{k}-space is fully observed. 

We initialize the process with an input image resulting from the observation of 10 measurements ($7.8\%$ kMA), containing only low-frequency information. The active acquisition process is depicted in Figure \ref{fig:simulation}, which contains qualitative intermediate results at different kMA values (including reconstructions, error maps, uncertainty estimates and acquisitions trajectories) as well as the progression of the mean uncertainty score and MSE on the test set. As shown in the figure, as we introduce additional measurements, the reconstruction quality improves and the error and uncertainty decrease; reaching very low values around kMA = $30\%$. Note that the uncertainty is condensed in complex image regions, often containing high frequency information. Moreover, higher uncertainty regions appear to have higher reconstruction error values. Please refer to the supplementary video for more simulation results.

\noindent
\textbf{Comparison to standard active acquisition heuristics.} 
We compare our evaluator-based approach to several baselines, including:
\vspace{-.1cm}
\begin{compactitem}
    \item Random+Copy(C): We randomly select an unobserved measurement, add it to the acquisition trajectory and compute the zero-filled reconstruction. We repeat this selection process without replacement until \textit{k}-space is fully observed.
    \item Random+C+Reconstruction(R): Following Random+C selection strategy, we pass the zero-filled solution through our reconstruction network every time a measurement is added.
    \item Order+C: We select measurements following the low to high frequency order. Following the copy strategy, we add the measurement to the acquisition trajectory and compute the zero-filled reconstruction. We repeat this selection process until \textit{k}-space is fully observed.
    \item Order+C+R: Following Order+C selection strategy, we pass the zero-filled solution through our reconstruction network  every time a measurement is added.
\end{compactitem} 
\vspace{-.1cm}

Figure \ref{fig:perf} analyzes the MSE as function of kMA. We observe that all methods have the same initial MSE and end up with zero MSE when all measurements are acquired. Random+C+R outperforms random+C notably, highlighting the benefit of applying the reconstruction network. However, order+C (even without any reconstruction) performs on par with random+C+R. This is not surprising, given that low frequency contains most of the information needed to reduce MSE. Finally, our method exhibits higher measurement efficiency when compared to the baselines.

\noindent
\textbf{ImageNet simulation.} MSE is unable to reflect how well the semantic details, which may be required for diagnosis, are recovered. Since we don't have access to classification information on our knee dataset, we reach out to an auxiliary classification dataset and test our pipeline. We evaluate the method by means of MSE and top-\emph{k} classification accuracy. Results are presented in Figure \ref{fig:perf_imgnet}. The MSE results for different acquisition heuristics follow the same pattern as in the knee dataset. Interestingly, when it comes to the classification accuracy, random+C+R outperform other baselines (which were better in terms of MSE, e. g. order+C+R), achieving results comparable to our method. This experiment suggests that semantic information could exist in arbitrary high-frequency parts of images. Our method demonstrates excellent effectiveness at recovering both image quality and semantic details.

\noindent
\textbf{Evaluator ablation study.} 
Finally, we compare our evaluator training strategy, described in subsection \ref{ssec:evaluator}, with two alternatives. First, we train our evaluator network with binary labels (following adversarial training of image-to-image translation networks \cite{isola2017image}), i.e. $0$ for spectral maps corresponding to unobserved measurements (fake), and $1$ for spectral maps corresponding to observed measurements (real). Second, we adapt the recently proposed \cite{devries2018leveraging} to score our spectral maps in terms of MSE. This approach trains a regression network on top of pre-trained regression model. Note that this is different to adversarial training, since the regression network does not affect the weights of the reconstruction network. The results of the comparison are shown in Figure \ref{fig:evaluator}, where the scores of different evaluators are depicted as a function of kMA. Note that only the scores of spectral maps corresponding to unobserved measurements are considered. A good evaluator should produce increasing scores (up to a maximum value of 1) as the number of acquired measurements increases. Similarly, the evaluator score variance should decrease with the number of acquired measurements. As it can been observed, our method is the only one satisfying both requirements, highlighting the benefits of our evaluator design.

\vspace{-.35cm}
\section{Conclusions}
\vspace{-.2cm}
In this paper, we presented a novel active acquisition pipeline for undersampled MRI reconstruction, which can iteratively suggests \emph{k}-space trajectories to best reduce uncertainty. We extensively validated our approach on a large scale knee dataset as well as on ImageNet, showing that (1) our evaluator design consistently outperforms alternative active acquisition heuristics; (2) our uncertainty estimates correlate with the reconstruction error and thus, can be used to trigger the halt signal of active acquisition at inference time; (3) our reconstruction architecture surpasses previously introduced architectures. Finally, we argued that the proposed method paves the road towards more applicable solutions for accelerating MRI, which ensure the optimal acquisition speedup while maintaining high fidelity image reconstructions with low uncertainty. 

\noindent
{\textbf{Acknowledgements}: We would like to thank Jure Zbontar, Anuroop Sriram, Nafissa Yakubova, Mike Rabbat, Erich Owens, Larry Zitnick, Florian Knoll, Jakob Assl{\"a}nder, Daniel K. Sodickson and everyone in the fastMRI team for their support and discussions. Finally, we extend our gratitute to Nicolas Ballas, Amaia Salvador, Lluis Castrejon and Joelle Pineau for their helpful comments.}

{\small
\bibliographystyle{ieee}
\bibliography{egbib}
}

\newpage
\setcounter{section}{0}
\section*{Supplementary Material}

\begin{figure*}[h]
    \centering
    \includegraphics[width=0.8\textwidth]{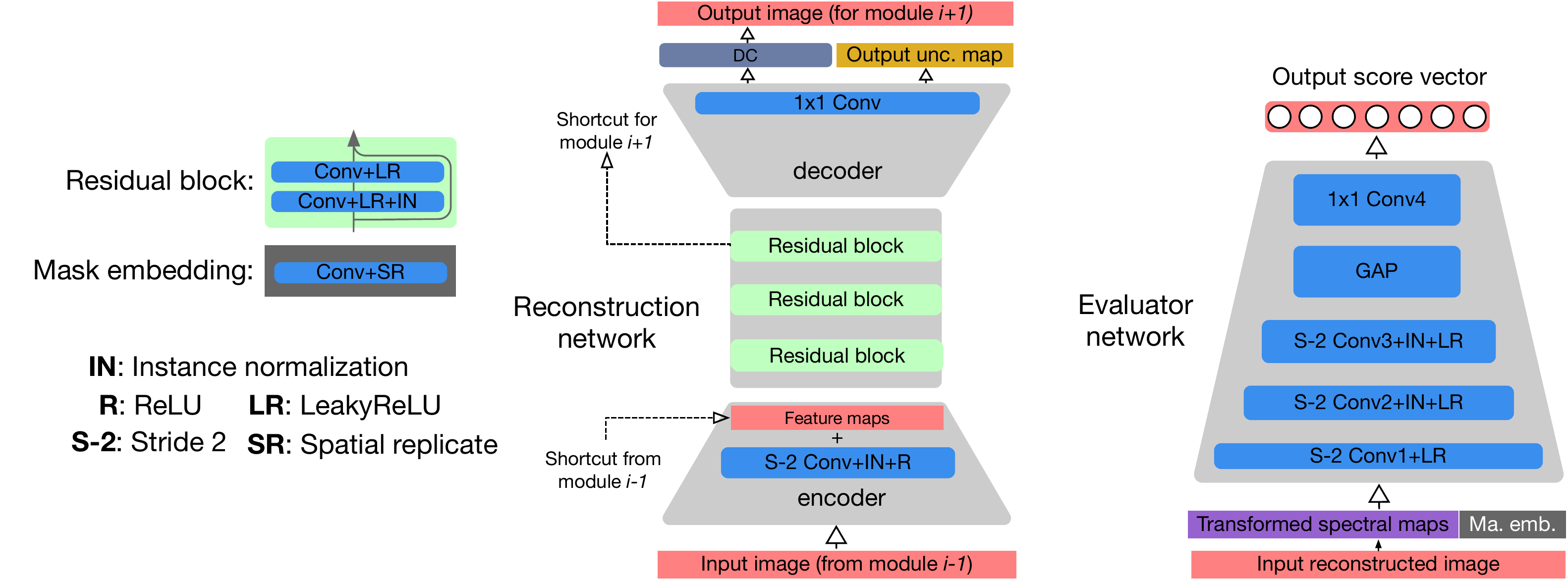}
    \vspace{0.2cm}
    \caption{Illustration of the reconstruction (left) and evaluator (right) networks. } \label{fig:arch} 
\end{figure*}

\begin{table*}[h]
\centering
\begin{tabularx}{0.70\textwidth}{lc cc}
\toprule
& Type & Output size & Comments  \\
\midrule
Input image   & - &  $2\times128\times128$  &   \\ 
\multirow{3}{*}{Encoder} & $Conv(2,3,1)$ & $128\times64\times64$ & \\
        & $Conv(2,3,1)$ & $256\times32\times32$ &  \\
        & $Conv(2,3,1)$ & $512\times16\times16$ & Skip-add from module \textit{i-1} \\ \addlinespace
        
\multirow{3}{*}{Residual blocks} & $ResBlock_1$ & $512\times16\times16$ & \\
                                & $ResBlock_2$ & $512\times16\times16$  &\\
                                & $ResBlock_3$ & $512\times16\times16$ & Skip to module \textit{i+1}  \\\addlinespace
    
\multirow{4}{*}{Decoder}  & $DeConv(2,4,1)$  & $256\times32\times32$ & \\
    & $DeConv(2,4,1)$ & $128\times64\times64$ & \\
    & $DeConv(2,4,1)$ & $64\times128\times128$ &   \\
    & $Conv(1,1,0)$ & $3\times128\times128$ & A channel for uncertainty  \\\addlinespace
DC  & -  & $2\times128\times128$ &  \\ \addlinespace
Output image  & -  & $2\times128\times128$ & Input to module \textit{i+1}\\
Output unc. map  & -  & $1\times128\times128$ & \\ \hline\hline 

\multirow{3}{*}{ResBlock}& Input & $512\times16\times16$ &  \\
                        & $Conv(1,3,1)$ & $512\times16\times16$ &  \\
                        & $Conv(1,3,1)$ & $512\times16\times16$ & Add the input \\
\bottomrule
\end{tabularx}

\vspace{0.4cm}
\caption{Details of the reconstruction network. In the table, we describe a single FC-ResNet module, which is composed of an encoder, three residual blocks, a decoder and a data consistency (DC) layer. In our reconstruction network, we repeat this module three times. In addition, to enhance the information flow between consecutive modules, we add a shortcut connection to directly link residual blocks between adjacent modules. Hence, each module can re-use the representations learned by its predecessor and enhance them with further computation. $Conv(2,3,1)$ denotes a convolutional layer with stride $2$, $3\times3$ kernel size, and reflection padding of $1$, while $DeConv(2,4,1)$ denotes a de-convolutional (transposed convolutional) layer with stride $2$, $4\times4$ kernel size, and reflection padding $1$. Each (de-)convolutional layer (except the last one) is followed by an instance normalization layer and ReLU. } \label{tab:rec} 
\end{table*}

\begin{table*}[h]
\centering
\begin{tabularx}{0.70\textwidth}{lc cc}
\toprule
& Type & Output size & Comments  \\
\midrule
Input spectral maps   & - &  $128\times128\times128$  & \\
Input mask column   & $Conv(1,1,1)$ &  $6\times1\times1$  & Replicate \& concatenate  \\ 
Input tensor  & - &  $134\times128\times128$  & \\ \addlinespace
\multirow{5}{*}{Evaluator} & $Conv(2,3,1)$ & $256\times64\times64$ & \\
    & $Conv(2,3,1)$ & $512\times32\times32$ &  \\
    & $Conv(2,3,1)$ & $1024\times16\times16$ & \\ 
    & $GAP$ & $1024\times1\times1$ & \\ 
    & $Conv(1,1,0)$ & $128\times1\times1$ & \\ \addlinespace
Output vector  & - &  $128$  & \\ 
\bottomrule
\end{tabularx}
\vspace{0.4cm}
\caption{Details of the evaluator network. In the first step, we create spectral maps. Then, since sampling trajectories have identical columns, we embed a single column into a $6$-d vector with a $1\times1$ convolutional layer and replicate the vector over all spatial locations of spectral maps. We follow the notation used to describe reconstruction network. Each convolutional layer (except the first one) is followed by an instance normalization layer and LeakyReLU with slope $0.2$. $GAP$ denotes global average pooling.} \label{tab:evaluator} 
\end{table*}

In the supplementary material, we first describe the details of both the reconstruction and the evaluator networks used in our system. Then, we explain how we ensure that our experimental results are not affected by simulating \emph{k}-space data from DICOM images, where the conjugate symmetry is present. Next, we discuss the inference time requirement for active acquisition. Finally, we introduce a video where we present additional results of our active acquisition system.

\section{Network Architectures}
Detailed diagrams of our reconstruction network as well as our evaluator network are shown in Figure \ref{fig:arch}. Moreover, in Table \ref{tab:rec}, we provide a description of all the building blocks of our reconstruction network. Similarly, in Table \ref{tab:evaluator}, we define the necessary components to replicate the design of our evaluator network.

\section{Conjugate Symmetry in DICOM data}
As mentioned in Section 5 of the paper, we use DICOM MRI images, which only store the image magnitude (i.e. $abs(\mathbf{x})$). We simulate the corresponding \textit{k}-space data $\mathbf{y}$ as $\mathbf{y} = \mathcal{F}(abs(\mathbf{x}))$. Since $abs(\mathbf{x}) \in \mathbb{R}^{N\times N}$, the complex valued matrix $\mathbf{y} \in \mathbb{C}^{N\times N}$ has the conjugate symmetric property \cite{rippel2015spectral,szeliski2011computer}. More precisely, each row $i$ in the top half of $\mathbf{y}$ has a conjugate symmetric row in the bottom half of $\mathbf{y}$:
\begin{equation}
    y_i = y^*_{(N-i)}, \quad i = 0,...,N{-}1,
\end{equation}
where $*$ denotes conjugate symmetry and $N$ denotes number of rows in $\mathbf{y}$. What follows is that the top half rows of the \textit{k}-space data $\mathbf{y}$ already contain all the frequency information needed to recover the corresponding image $abs(\mathbf{x})$.

Therefore, when simulating active acquisition in scenarios where \emph{k}-space is obtained from DICOM images, the conjugate symmetry of the data should be taken into account, since additional measurements could carry no further information. To deal with this, in all our experiments (including baselines), we make sure that when a \textit{k}-space row is selected, its conjugate symmetric row is also selected automatically. In this way, our system needs maximally $64$ iterations to fully observe the $128\times128$ \textit{k}-space. Please note that our strategy to select measurements in the \textit{k}-space only affects the cardinally of the selection process and does not make the proposed approach nor the baselines less generalizable.

\section{Inference time}
Inference time is an important factor to guarantee the applicability of active acquisition algorithms. We use the MRI scanner protocol of our data acquisition as an example to illustrate the time requirement.

There are many details that lead to the MRI scan time, the first and foremost the pulse sequence. Before being stored in PACS (picture archiving and communication system), our data were acquired with a 2D turbo-spin echo sequence (TSE). TSE sequences operate by acquiring a small batch of data (4 lines of k-space in this case) from one 2D slice, then repeating this for all the other slices. Considering refocusing pulses, it takes about $7$ ms to acquire a single line. Finally, the repetition time is about $2.7$ s. In summary, inference speed for single line selection would need to be roughly $7$ ms, whereas inference speed for batch selection would need to be $2.7$ s. Initial (non-optimized) implementation of our pipeline has inference time in order of $15$ ms on a single GPU.

\end{document}